\documentclass[conference]{IEEEtran}
\IEEEoverridecommandlockouts
\usepackage{cite}
\usepackage{amsmath,amssymb,amsfonts}
\usepackage{algorithmic}
\usepackage{graphicx}
\usepackage{textcomp}
\usepackage{multirow} 
\usepackage{booktabs}

\usepackage{xcolor}
\usepackage{url}

\def\BibTeX{{\rm B\kern-.05em{\sc i\kern-.025em b}\kern-.08em
    T\kern-.1667em\lower.7ex\hbox{E}\kern-.125emX}}
\begin{document}

\title{Restoring Real-World Images with an Internal Detail Enhancement Diffusion Model}


\author{
  Peng Xiao, \quad
  Hongbo Zhao, \quad
  Yijun Wang, \quad
  Jianxin Lin \\
  Hunan University, China \\
  \texttt{\{napping, hongbozhao, wyjun, linjianxin\}@hnu.edu.cn}
}

\maketitle

\begin{abstract}
Restoring real-world degraded images, such as old photographs or low-resolution images, presents a significant challenge due to the complex, mixed degradations they exhibit, such as scratches, color fading, and noise. Recent data-driven approaches have struggled with two main challenges: achieving high-fidelity restoration and providing object-level control over colorization. While diffusion models have shown promise in generating high-quality images with specific controls, they often fail to fully preserve image details during restoration. In this work, we propose an internal detail-preserving diffusion model for high-fidelity restoration of real-world degraded images. Our method utilizes a pre-trained Stable Diffusion model as a generative prior, eliminating the need to train a model from scratch. Central to our approach is the Internal Image Detail Enhancement (IIDE) technique, which directs the diffusion model to preserve essential structural and textural information while mitigating degradation effects. The process starts by mapping the input image into a latent space, where we inject the diffusion denoising process with degradation operations that simulate the effects of various degradation factors. Extensive experiments demonstrate that our method significantly outperforms state-of-the-art models in both qualitative assessments and perceptual quantitative evaluations. Additionally, our approach supports text-guided restoration, enabling object-level colorization control that mimics the expertise of professional photo editing.
\end{abstract}


\section{Introduction}
The task of restoring the visual artifacts in real-world degraded images, such as old photographs or low-resolution images covered by a variety of distortions, remains a challenging research area yet not well resolved. Despite recent advancements in data-driven methodologies  \cite{wan2020bringing,xu2022pikfix,wang2018esrgan,liang2021swinir}, this field still grapples with two main challenges. First, there is a critical need to produce high-fidelity restored images with vivid colors and photorealistic details. Second, achieving precise control over object-level color nuances remains an unsolved task for old photo restoration. Notably, the emergence of state-of-the-art diffusion models (DMs) \cite{ho2020denoising, rombach2022highresolution,zhang2023adding} capable of generating exceptional-quality images with predefined attributes. However, these models still struggle in their quest to faithfully retain the intricate details given the low-quality images due to their stochastic nature.

To address the aforementioned challenge, a common practice is to train an image restoration model from scratch  \cite{saharia2022superresolution,lugmayr2022repaint,whang2022deblurring}. To maintain the image details, these methods usually take the low-quality image as an additional input to constrain the output space. While such approaches have achieved success on tasks such as image super-resolution \cite{saharia2022superresolution} and image deblurring  \cite{whang2022deblurring}, the design of these approaches often focuses on one certain image degradation and starts the training from scratch, resulting in limited generalizability. Meanwhile, large-scale diffusion models \cite{rombach2022highresolution,zhang2023adding} for image generation or text-to-image generation have exhibited superior performance in generating high-quality images. Therefore, alternative approaches \cite{choi2021ilvr,fei2023generative,wang2024exploiting} take advantage of such generative prior for image restoration by introducing constraints into the reverse diffusion process. The design of these constraints also requires prior knowledge of the image degradations and an optimization process for every single image, limiting its feasibility in practice. Therefore, few existing works based on diffusion models address the problem of real-world degraded image restoration with multiple unknown degradations, especially for old photo restoration.  

\begin{figure}[!t]
\centering
  \includegraphics[width=0.48\textwidth]{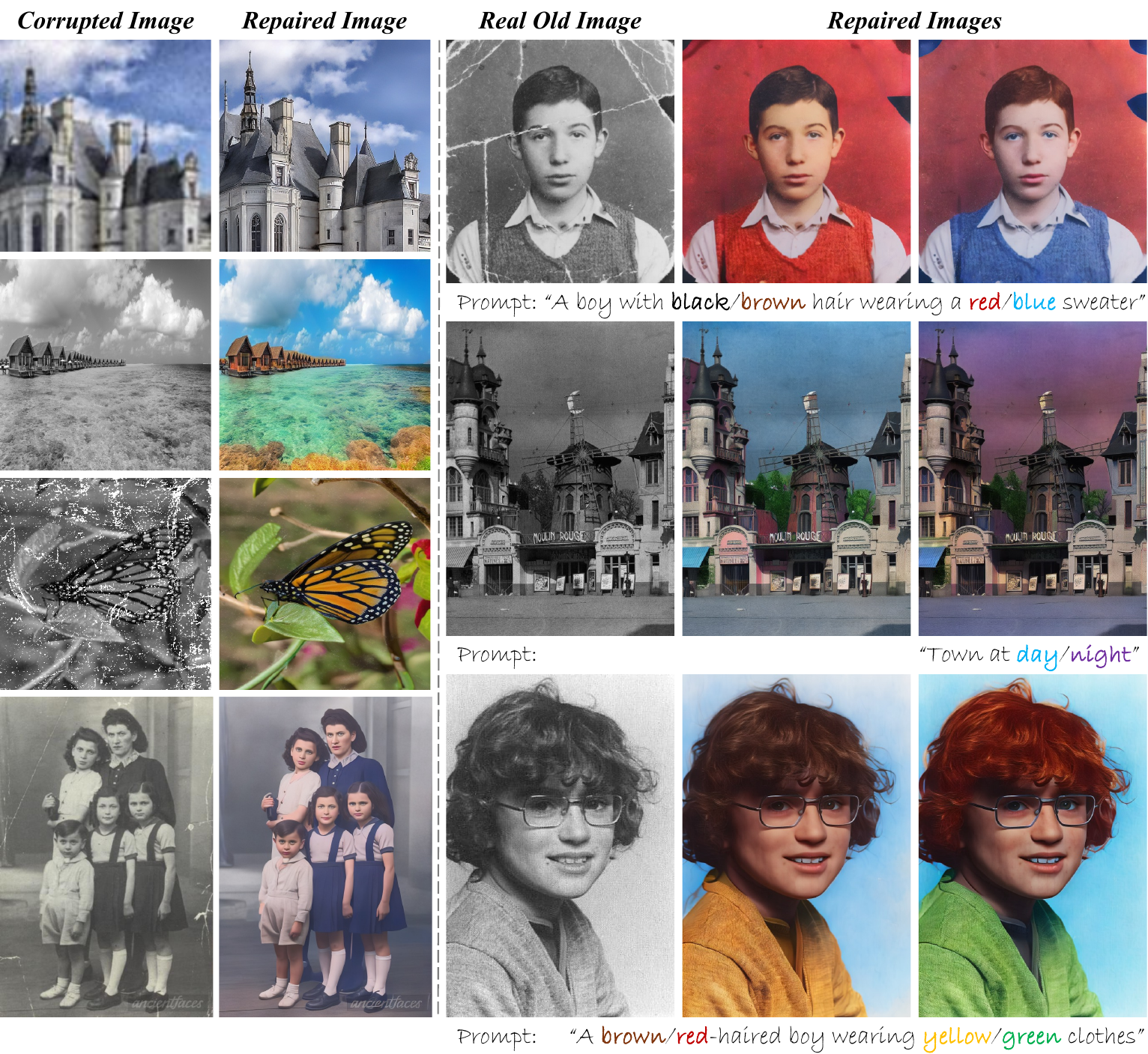}
  \caption{\textbf{Image Restoration under Various Real-World Degradations} Given a degraded image with low quality, our method produces high-fidelity restorations and enables \textbf{object-level color control} when provided with text prompts.}
  \label{fig:title_page}
\end{figure}

In this paper, our approach seeks to incorporate the advantages of the large-scale diffusion model's generative prior knowledge for low-quality image restoration with unknown degradations, and preserve the intuitive text-based editing abilities at the same time. To this end, we introduce an novel approach aiming to achieve high-fidelity text-guided image restoration using internal detial-preserving diffusion models. Essentially, we introduce Internal Image Detail Enhancement (IIDE) as a fine-tuning technique, aiming to direct the generative process within the diffusion model. Its objective is to produce high-quality images from low-quality input while meticulously preserving the image's intricate details. IIDE introduces constraints on the diffusion process, ensuring that the restored high-quality image remains faithful to the content, regardless of the specific low-quality conditions. Specifically, IIDE utilizes the Denoising Diffusion Implicit Model (DDIM)  \cite{song2020denoising} to automatically estimate a degraded version of the high-quality image. This approach alleviates limitations related to manual degradation design and guarantees the preservation of image details, thus enhancing the overall quality of the restored images. It is worth noting that our method fine-tunes a frozen pre-trained diffusion model with a limited number of trainable parameters, eliminating the need to train the diffusion model from scratch.

As demonstrated in Fig. \ref{fig:title_page}, our method can successfully produce high-fidelity restored images with vivid colors and photorealistic details conditioned on low-quality images with multiple unknown degradations. 
Furthermore, our approach enables user control in old photo restoration, allowing for precise adjustments to the semantic similarity of the restored image based on textual prompts, as shown in the right section of Fig. \ref{fig:title_page}. Extensive experiments conducted on synthetic and real-world datasets demonstrate that our method provides effective object-level control over diversity while preserving high visual consistency, revealing its superiority over prior state-of-the-art models.

Our contributions can be summarized as below:

\begin{itemize}
    \item The paper introduces a novel approach that addresses the challenge of text-guided image restoration with multiple unknown degradations using diffusion priors. 
    
    \item An Internal Image Detail Enhancement (IIDE) method is proposed to ensure the generation of detail-preserved images within the diffusion model training process.

    \item Extensive experiments verify that our method outperforms state-of-the-art methods on real-world image restoration, and enables object-level color control especially for old photo restoration, which has not been explored before. 
\end{itemize}

\section{Related Works}

\subsection{Traditional Image Restoration}
Pioneer works\cite{dong2015image, dong2015compression} based on Convolution Neural Network (CNN) has achieved impressive performance on Image Restoration (IR) tasks. Recently, Transformer\cite{vaswani2017attention} has gained much popularity in the computer vision community. Compared with CNN, transformers can model global interactions between different regions and achieve better performance on IR tasks\cite{chen2021pre, liang2021swinir, zamir2022restormer}.

\subsection{Image Restoration with Diffusion Model}
Diffusion models (DMs) have disrupted the IR field, and further closed the gap between image quality and human perceptual preferences compared to previous generative methods, i.e., GAN\cite{goodfellow2014generative}. Priors knowledge from pre-trained DMs are proven to be greatly helpful in most IR tasks, like image colorization\cite{carrillo2023diffusart}, single image super-resolution\cite{wang2024exploiting, wang2024sinsr, xia2023diffir} and deblurring\cite{ren2023multiscale}. 


\begin{figure*}[!t]
    \centering

        \includegraphics[width=0.85\textwidth]{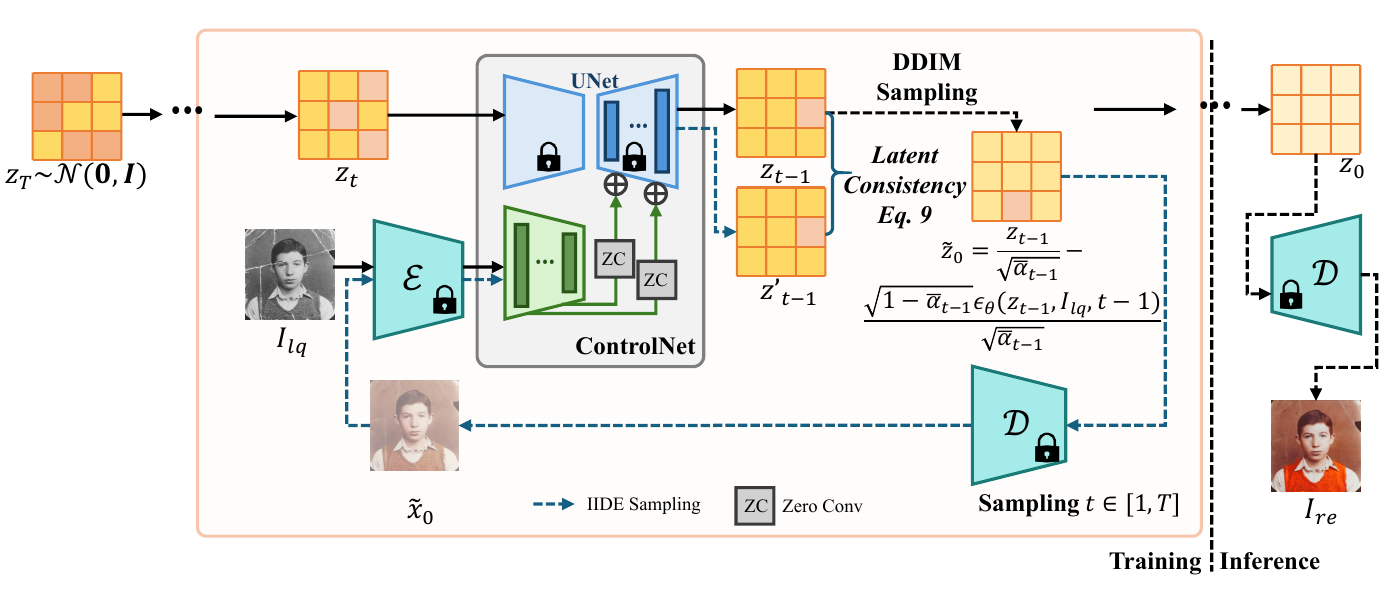}
    
    \caption{The framework of our proposed method, which is obtained by finetuning a pre-trained diffusion model with the Internal Image Detail Enhancement (IIDE) mechanism. IIDE constructs a self-regularization that enforces the denoised result to keep the maximum image details from the image condition.}
    \label{fig:framework}
\end{figure*}

\section{Method}
In this study, we seek to incorporate the advantages of a well-trained diffusion model's prior for low-quality image restoration and preserve the intuitive text-based editing abilities at the same time. In this section, we begin by introducing the foundational DDIM method as the prerequisite knowledge for Internal Image Detail Enhancement (IIDE). Following that, we provide a comprehensive explanation of our IIDE fine-tuning strategy, which optimizes the backward generative process. Lastly, we delve into the implementation for enhancing the stability of translated outcomes within the diffusion model.
The overall framework of our proposed method is exihibited in Fig. \ref{fig:framework}.

\subsection{Background and Preliminaries}
In this paper, we implement our method based on the large-scale text-to-image latent diffusion model, Stable Diffusion \cite{rombach2022highresolution}. Diffusion models learn to generate data samples through a sequence of denoising steps that estimate the score of the data distribution. In order to achieve better efficiency and stabilized training, Stable Diffusion pretrains an autoencoder that converts an image $x$ into a latent $z_0=\mathcal{E}(x)$ with encoder $\mathcal{E}$ and reconstructs it with decoder $\mathcal{D}$. The diffusion and denoising processes are performed in the latent space. In the diffusion process, Gaussian noise with variance $\beta_t \in (0, 1)$ at time $t$ is added to the encoded latent $z_0 = \mathcal{E}(x)$ for producing the noisy latent:

\vspace{-0.5em} 
\begin{equation}\label{eq:z02zt}
z_t = \sqrt{\bar{\alpha}_t}z_0 + \sqrt{1 - \bar{\alpha}_t}\epsilon, 
\end{equation}

\noindent where $t \in \{1, \ldots, T\}$, $\epsilon \sim \mathcal{N}(0, I)$, $\alpha_t = 1 - \beta_t$, and $\bar{\alpha}_t = \prod_{i=0}^{t}\alpha_i$.

In the training phase, the model with $\theta$ as parameter is trained to predict this noise $\epsilon$ from the latent variables $z_t$. In text-guided diffusion models, the model is further conditioned by a feature representation (an embedding) $C$, which is derived from a text prompt $P$, often obtained using a text encoder such as CLIP \cite{radford2021learning}.

The loss function for training the model is defined as the Mean Squared Error (MSE) between the predicted noise $\epsilon_\theta$ and the actual noise $\epsilon$:
\vspace{-0.5em} 
\begin{equation}\label{eq:ddpm_loss}
L(\theta) = \mathbb{E}_{t \sim U(1,T), \epsilon \sim \mathcal{N}(0,I)} \lVert \epsilon - \epsilon_\theta(z_t, t, C) \rVert^2_2,  
\end{equation}

\noindent where $U(1, T)$ represents the uniform distribution over the set $\{1, \ldots, T\}$, and $\mathcal{N}(\mu, \Sigma)$ denotes the multivariate Gaussian distribution with mean $\mu$ and covariance $\Sigma$. In the inference stage, a sample $\tilde{x}_0=\mathcal{D}(\tilde{z}_0)$ is generated by passing the generated representation $\tilde{z}_0$ through the decoder $\mathcal{D}$.

While the reverse diffusion process in denoising diffusion probabilistic models (DDPMs)  \cite{ho2020denoising} is inherently stochastic, the reverse process employed by the DDIM sampling method  \cite{song2020denoising} becomes deterministic while still generating the same data distribution. Hence, in the subsequent discussion, we adopt DDIM for the sampling method. The DDIM computes the latent variable \(z_{t-1}\) at diffusion step \(t-1\) from the latent variable \(z_t\) at step \(t\) using the formula:

\vspace{-0.5em} 
\begin{equation}
\resizebox{.9\hsize}{!}{$
\begin{aligned}\label{eq:zt-1}
z_{t-1} &= \sqrt{\frac{\alpha_{t-1}}{\alpha_t}} z_t \\
&\quad + \alpha_{t-1} \left( \sqrt{\frac{1 - \alpha_{t-1}}{\alpha_{t-1}}} - \sqrt{\frac{1 - \alpha_t}{\alpha_t}} \right) \epsilon_\theta(z_t, t, C)
\end{aligned}
$}
\end{equation}

\noindent where \(\alpha := (\alpha_1, \ldots, \alpha_T) \in \mathbb{R}_{\geq 0}^T\) represents the hyper-parameters that determine noise scales at \(T\) diffusion steps.

Specifically, in the sampling process, DDIM can estimate a clean image representation $\tilde{z}_0$ directly from the noisy latent $z_t$ by estimating the noise in $z_t$  as follow:
\begin{equation}\label{eq:zt2z0}
\tilde{z}_0 = \frac{z_t}{ \sqrt{\bar{\alpha}_t}} - \frac{\sqrt{1 - \bar{\alpha}_t} \epsilon_\theta(z_t, t, C)}{\sqrt{\bar{\alpha}_t}}.
\end{equation}

Thus, we can obtain a clean image prediction $\tilde{x}_0 = \mathcal{D}(\tilde{z}_0)$ from an arbitrary noisy latent $z_t$.

\subsection{Internal Image Detail Enhancement (IIDE)}

With a low-quality image represented as $I_{lq}$, our goal is to restore $I_{lq}$ to get a high-quality image $I_{re}$ similar to the real high-quality image $I_{hq}$, while enabling both null text guidance and text guidance with target prompt $P$. Due to the inherent stochasticity of diffusion model, the main challenge of the restoration process is faithfully converting $I_{lq}$ to high-quality space while retaining intrinsic image details. Therefore, we propose Internal Image Detail Enhancement (IIDE) to explicitly constraint diffusion iterations to ensure the model preserves the image details conditioned on $I_{lq}$, as shown in Fig. \ref{fig:framework}.

Let $\phi_{\omega}(\cdot)$ denote an image degradation mixed operation, which involves various processes and factors that can negatively affect the quality of an image, such as noise, blurring, compression, or other forms of distortion while maintaining the content of the image. For example,  $I_{lq}$ is the result by applying a specific degradation operation $\phi_{\omega^{\ast}}(\cdot)$ on $I_{hq}$. In practical scenarios, it is observed that the degradation mixed operation $\phi_{\omega}(\cdot)$ can exhibit a wide range of quality levels, spanning from severe degradation to minimal degradation, and sometimes even being non-degradation. Thus, $I_{lq}$ should be enforced to ensure that the restored image $I_{re}$ conditioned on $I_{lq}$ should be equal to another restored image conditioned on $\phi_{\omega}(I_{hq})$:

\vspace{-0.5em} 
\begin{equation}
    I_{re}=G(I_{lq},C)=G(\phi_{\omega}(I_{hq}),C),
\end{equation}
where $C$ could be the embedding of content prompt $P$ or a null text $\emptyset$, and $G$ is the restoration model. Utilizing the forward process \(q(z_{t} | z_0)\) of Eq. \ref{eq:z02zt}, where $z_0=\mathcal{E}(I_{hq})$, for step $t$, the model has each transition that obeys the Markov transition rule and  follows criterion as below:
\begin{equation}
\label{eq:condreverse}
p_{\theta}(z_{t-1} | z_{t}, C, I_{lq}) \approx p_{\theta}(z_{t-1} | z_{t}, C, \phi_{\omega}(I_{hq}) ).
\end{equation}

Although such constraints on the diffusion model explicitly make the requirement that a $G$ should learn to generate the high-quality image with the same image content regardless of different low-quality conditions, it still has two main shortcomings: 1) 
manual design of $\phi_{\omega}(\cdot)$ can not approximate real mixed degradations perfectly; 2) the image details in $I_{lq}$ tranferred to $z_{t-1}$ has still not been guaranteed to be preserved.

Reminding that, given the low-quality image $I_{lq}$, our target is to maximize the likelihood $p_{\theta}(z_0|I_{lq})$ that is  equivalent to:

\begin{equation}\label{eq:p_theta}
    p_{\theta}(z_0|I_{lq}) = \int p_{\theta}(z_0|\tilde{x}_0)p_{\theta}(\tilde{x}_0|I_{lq}) d\tilde{x}_0,
\end{equation}

\noindent where $ p_{\theta}(z_0|\tilde{x}_0)$ can be assumed to restore image with the new condition $\tilde{x}_0$ and  $p_{\theta}(\tilde{x}_0|I_{lq})$ can naturally be rewritten as the diffusion denoising process:

\vspace{-0.5em} 
\begin{equation}\label{eq:p_theta2}
\resizebox{.9\hsize}{!}{$
p_{\theta}(\tilde{x}_0|I_{lq}) = \int \int p(\tilde{x}_0|z_{t-1})p_{\theta}(z_{t-1}|z_{t})q(z_{t}|I_{lq}) dz_{t-1} dz_{t},
$}
\end{equation}

\noindent where $p(\tilde{x}_0|z_{t-1})$ and $p_{\theta}(z_{t-1}|z_{t})$ are defined at Eq. \ref{eq:zt2z0} and Eq. \ref{eq:zt-1} respectively. As paired $x$ (or $I_{hq}$) and $I_{lq}$ share similar content, $q(z_{t}|I_{lq}) \approx q(z_{t}|x)$ when $t$ is large enough, and $q(z_{t}|x)$ is defined at Eq. \ref{eq:z02zt}. Thus, we propose the IIDE process to be integrated into a diffusion model for image restoration according to Eq. \ref{eq:p_theta} and Eq. \ref{eq:p_theta2}.

Specifically, the IIDE proposes to estimate a $ \tilde{x}_0=\mathcal{D}(\tilde{z}_0)$ as a degradation version of $I_{hq}$ derived from $z_{t-1}$ defined by Eq. \ref{eq:zt2z0}, and we can rewrite Eq. \ref{eq:condreverse} as:
\begin{equation}
\label{eq:new_condreverse}
p_{\theta}(z_{t-1} | z_{t}, C, I_{lq}) \approx p_{\theta}(z_{t-1} | z_{t}, C, \tilde{x}_0 ).
\end{equation}

\begin{figure}[!t]
  \centering
  \includegraphics[width=0.49\textwidth]{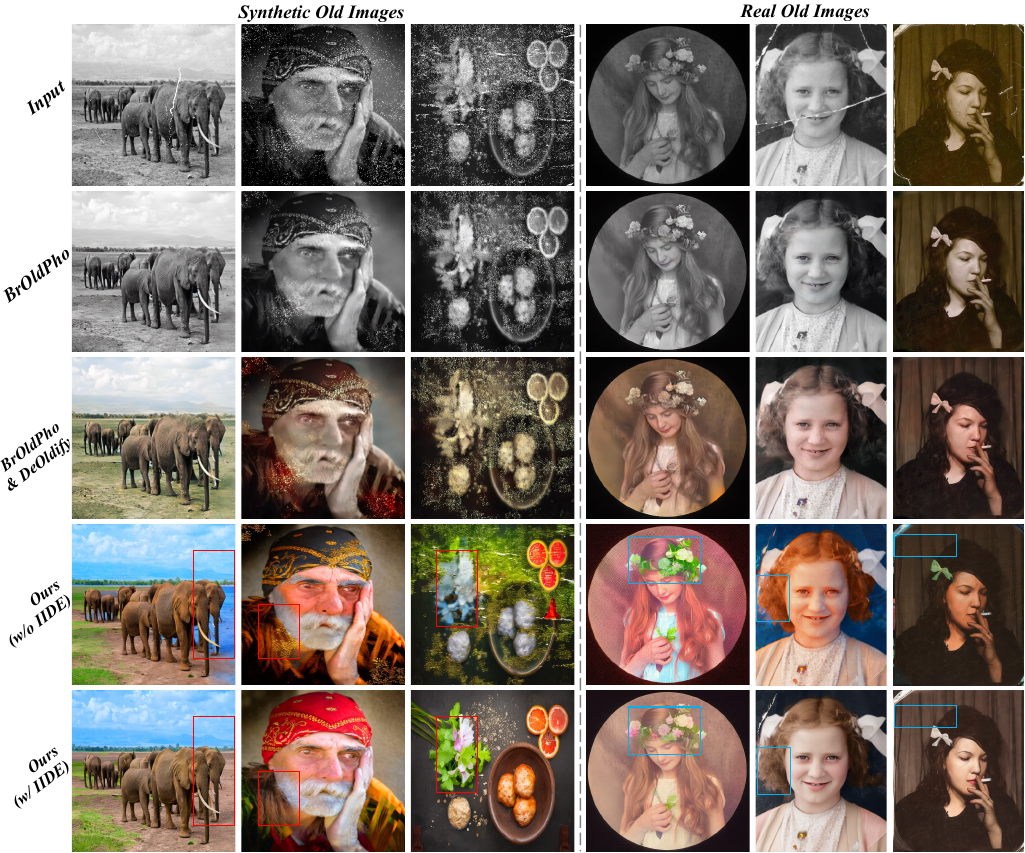}
  \caption{Visual comparison of old photo restoration methods. Zoom-in for better details.}
  \label{fig:sync and real old}
\end{figure}

Thus, the two main shortcomings have been alleviated because 1) the new degraded image $ \tilde{x}_0$ is automatically produced by the diffusion model at different step $t$; 2) $z_{t-1}$ constructs a self-regularization that two $z_{t-1}$s respectively generated by condition $I_{lq}$ and $\tilde{x}_0$ (predicted by $z_{t-1}$ ) should keep the maximum similarity, meaning $z_{t-1}$ itself should preserve image details from $I_{lq}$. The results of applying the IIDE process can be found in Fig. \ref{fig:sync and real old}, where we can see that IIDE effectively addresses the instability problem of the diffusion model.

\subsection{Diffusion Training with IIDE}

In practice, we employ a mix-up training technique to jointly train the model with two kinds of conditional settings. Specifically, for the probability of ``$p_{iide}$'' we configure the context information, denoted as ``$C_I$'', with the original image condition $I_{lq}$ to train the model. For all other cases, we use the new image condition $\tilde{x}_0$ that is derived from $z_{t-1}$ produced by $z_{t}$.

We freeze all the parameters in the Stable Diffusion model, and only train the added module ControlNet \cite{zhang2023adding}, parameterized by $\theta$. We follow the diffusion model training loss as Eq. \ref{eq:ddpm_loss} to optimize our model. 

\section{Experiments}

\subsection{Implementation Details}
We present a novel method which is fine-tuned on Stable Diffusion 2.1\textsuperscript{{}}\footnote{\url{https://huggingface.co/stabilityai/stable-diffusion-2-1-base}} on two unique task types based on the architectural scheme of ControlNet \cite{zhang2023adding}:

\subsubsection{Old Photo Restoration} Data used in Old Photo Restoration comprises a complex amalgamation of structured and unstructured degradation. To tackle this, we augment our model with an additional condition, i.e., a scratch mask derived from $I_{lq}$, integrated with degraded images. 

\subsubsection{Image Super-Resolution} In this task, we retain $I_{lq}$ as the sole vision condition and set text prompts to null during both the training and inferencing stages. Similar to StableSR \cite{wang2023exploiting}, we introduce a controllable feature transformation module \cite{zhou2022towards} on the sampled latent codes after the training stage, which aims to achieve a tradeoff between quality and fidelity of image restoration.

We designate $p_{iide}$ as $0.5$ for both tasks to ensure a steadier training process. This provides an equal probability of substituting the original vision condition $I_{lq}$ with $\tilde{z}_0$. 

\subsection{Metrics}
To evaluate on both synthetic testing dataset with reference images and real-world datasets, we follow SinSR\cite{wang2024sinsr} utilize PSNR, SSIM, and LPIPS\cite{zhang2018effectiveness} to measure the fidelity performance, and use CLIPIQA\cite{wang2023exploring} and MUSIQ\cite{ke2021musiq} these two non-reference metrics to justify the realism of all the images.

\subsection{Compared Methods}
Due to space limitations, we put the detials of our compared methods to \textbf{Appendix A}. Please refer to the appendix for further details.

\subsection{Experimental Results}
\subsubsection{Evaluation on Text-Guided Image Colorization} 
Due to space limitations, we put the detials of our the evaluation of text-guided image colorization task to \textbf{Appendix B}. Please refer to the appendix for further analysis and visual comparasion.

\subsubsection{Evaluation on Old Photo Restoration}
We conducted a quantitative comparison using synthetic old photos from the DIV2K dataset\cite{agustsson2017ntire}. Our model was evaluated against the BrOldPho\cite{wan2020bringing} and DeOldify\cite{anctic2021deoldify} pipeline, revealing notable performance differences.

Tab. \ref{tab:old_photo} shows that, although our method has a slight reduction in PSNR and SSIM, indicating a minor trade-off in structural similarity, it outperforms both BrOldPho and the Br-DeOld combination in perceptual metrics, which are crucial for old photo restoration. Specifically, our model improves the CLIPIQA\cite{wang2023exploring} score by 53.6\% over BrOldPho, aligning more closely with human perception. The MUSIQ\cite{ke2021musiq} score increases by 29.5\% compared to Br-DeOld, demonstrating better capture of multi-scale image characteristics. Additionally, the FID metric is reduced by 18.4\% compared to Br-DeOld, indicating that our images more closely match real image distributions.

Qualitative results shown in Fig. \ref{fig:sync and real old} reveal that our method excels in restoring both synthetic and authentic old photos with superior detail and color accuracy. The Iterative Image Detail Enhancement (IIDE) technique effectively preserves fine details and improves color fidelity, as highlighted in the blue boxes in Fig. \ref{fig:sync and real old}.

In conclusion, while there is a slight trade-off in traditional metrics, our model exhibits superior performance in perceptual metrics and successfully replicates real image distributions, thereby validating its efficacy for old photo restoration tasks. For further visual comparisons of old photo restoration, we refer the reader to \textbf{Appendix C} where additional comparisons are provided.

\begin{table}[t]
    \centering
    \caption{Quantitative comparison of Old Photo Restoration on DIV2K.}
    \resizebox{\columnwidth}{!}{ 
    \small 
    \setlength{\tabcolsep}{4pt} 
    \renewcommand{\arraystretch}{1.2} 
    \begin{tabular}{c|cccccc}
        \toprule
        \multirow{2}{*}{\textbf{Methods}} & \multicolumn{6}{c}{\textbf{Metrics}} \\
        \cline{2-7}
        & \textbf{PSNR$\uparrow$} & \textbf{SSIM$\uparrow$} & \textbf{LPIPS$\downarrow$} & \textbf{FID$\downarrow$} & \textbf{CLIPIQA$\uparrow$} & \textbf{MUSIQ$\uparrow$} \\
        \midrule
        BrOldPho\cite{wan2020bringing} & \underline{29.73} & 0.8475 & 0.4946 & 112.6 & 0.4452 & 50.46 \\
        BrOldPho \& DeOldify\cite{anctic2021deoldify} & \textbf{30.20} & 0.8502 & \textbf{0.4597} & \underline{75.98} & 0.4251 & 54.70 \\
        Ours (w/o IIDE) & 26.81 & \underline{0.9181} & 0.5845 & 78.49 & \underline{0.6048} & \underline{61.70}\\
        Ours (w/ IIDE) & 29.11 & \textbf{0.9352} & \underline{0.4902} & \textbf{62.03} & \textbf{0.6837} & \textbf{70.84} \\
        \bottomrule
    \end{tabular}
    }
    \label{tab:old_photo}
\end{table}

\begin{figure}[!t]
  \centering
  \includegraphics[width=0.48\textwidth]{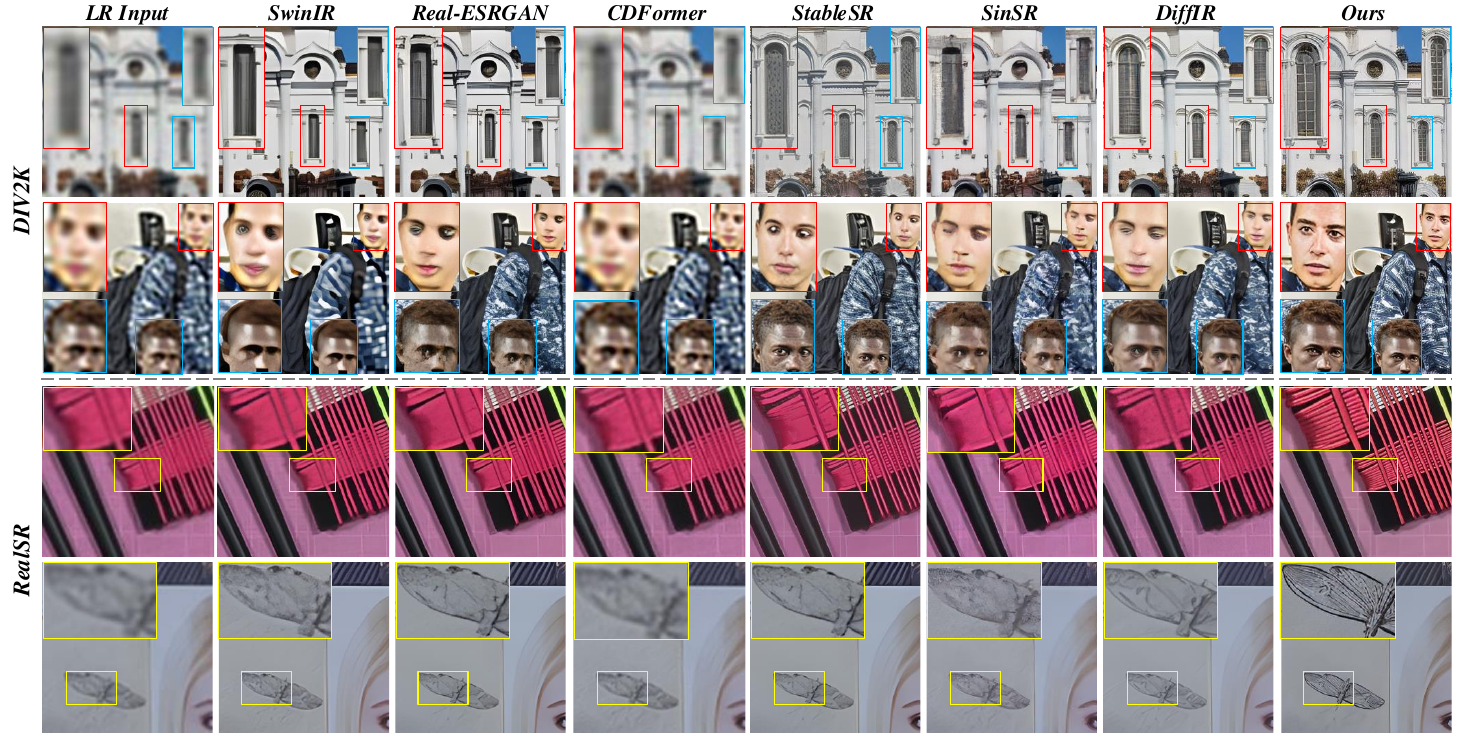}
  \caption{Visual comparisons of SR methods. Zoom-in for better details.}
  \label{fig:sr}
\end{figure}

\subsubsection{Image Super-Resolution} We evaluated our approach on both synthetic (DIV2K dataset \cite{agustsson2017ntire}) and real-world datasets (RealSR \cite{cai2019toward} and DRealSR \cite{wei2020component}). As shown in Tab. \ref{tab:sr_combined_div2k} and Tab. \ref{tab:sr_combined_realsr_drealsr}, our method consistently outperforms state-of-the-art SR methods in perceptual metrics (CLIPIQA and MUSIQ), which better align with human visual perception. On the DIV2K dataset, we achieve a 3.6\% improvement in CLIPIQA and 1.6\% in MUSIQ over StableSR \cite{wang2024exploiting}. On real-world datasets, our method demonstrates even greater advantages, with a 7.4\% increase in CLIPIQA and a slight 0.3\% gain in MUSIQ over StableSR on RealSR dataset \cite{cai2019toward}, and a 5.5\% and 7.3\% improvement in CLIPIQA and MUSIQ, respectively, on DRealSR dataset\cite{wei2020component}. These results underscore the robustness of our approach in generating perceptually accurate and visually pleasing super-resolved images, even in the presence of challenging real-world distortions.

It is important to acknowledge, however, that our method exhibits a slight trade-off in traditional quality metrics, such as PSNR and SSIM, where it falls behind methods like RealESRGAN \cite{wang2021realesrgan} and StableSR \cite{wang2024exploiting}. This discrepancy arises because PSNR and SSIM are primarily focused on pixel-level accuracy and structural similarity, which do not always correspond to the way humans perceive image quality. Our approach, in contrast, prioritizes perceptual metrics that emphasize fine-grained detail and natural visual aesthetics over pixel-perfect reconstruction. As such, while traditional metrics may not fully capture the advantages of our method, the perceptual gains are evident in the visual quality of the images, where our method produces more realistic and visually consistent results.

Fig. \ref{fig:sr} provides visual evidence of the strengths of our method. In the first row, our model successfully restores intricate architectural details, whereas methods like CDFormer \cite{liu2024cdformer} and SinSR \cite{wang2024sinsr} result in blurry or unnatural features. In the second row, our method delivers sharp, clear facial details, which are missing or poorly rendered in the results from other methods. This highlights our model’s superior ability to preserve structural fidelity and enhance image details, aligning closely with human visual expectations. For additional visual comparisons, readers are referred to \textbf{Appendix D}.
\begin{table}[!t]
    \centering
    \caption{Quantitative comparison of with the SOTA SR methods on synthetic DIV2K dataset.}
    \resizebox{\columnwidth}{!}{ 
    \small 
    \setlength{\tabcolsep}{3pt} 
    \renewcommand{\arraystretch}{1.2} 
    \begin{tabular}{c|ccccccc}
        \toprule
        \multirow{2}{*}{Methods} & \multicolumn{6}{c}{Metrics} \\
        \cline{2-7}
        & PSNR~$\uparrow$ & SSIM~$\uparrow$ & LPIPS~$\downarrow$ & FID~$\downarrow$ & CLIPIQA~$\uparrow$ & MUSIQ~$\uparrow$ \\
        \midrule
        SwinIR\cite{liang2021swinir} & 32.77 & 0.8169 & 0.3348 & 40.69 & 0.5276 & 59.06 \\
        RealESRGAN\cite{wang2021realesrgan} & \underline{33.10} & \underline{0.8207} & 0.3243 & 41.34 & 0.5309 & 60.35 \\
        CDFormer\cite{liu2024cdformer} & 32.82 & 0.8176 & 0.7042 & 72.19 & 0.3351 & 23.32 \\
        SinSR\cite{wang2024sinsr} & 32.62 & 0.7964 & 0.3316 & 38.36 & 0.6591 & 62.90 \\
        DiffIR\cite{xia2023diffir} & \textbf{33.46} & \textbf{0.8255} & \textbf{0.2475} & \textbf{25.38} & 0.5800 & 62.49 \\
        StableSR\cite{wang2024exploiting} & 32.08 & 0.7893 & \underline{0.3187} & \underline{26.61} & \underline{0.6804} & \underline{66.04} \\
        Ours & 31.84 & 0.7870 & 0.3495 & 31.13 & \textbf{0.7046} & \textbf{67.06} \\
        \bottomrule
    \end{tabular}
}
    \label{tab:sr_combined_div2k}
\end{table}

\begin{table}[!t]
    \centering
    \caption{Quantitative comparison of with the SOTA SR methods on real-world datasets.}
    \resizebox{\columnwidth}{!}{ 
    \small 
    \setlength{\tabcolsep}{3pt} 
    \renewcommand{\arraystretch}{1.2} 
    \begin{tabular}{c|cc|cc}
        \toprule
        \multirow{2}{*}{Methods} & \multicolumn{2}{c|}{RealSR Dataset} & \multicolumn{2}{c}{DRealSR Dataset} \\
        \cline{2-5}
        & CLIPIQA~$\uparrow$ & MUSIQ~$\uparrow$ & CLIPIQA~$\uparrow$ & MUSIQ~$\uparrow$ \\
        \midrule
        SwinIR\cite{liang2021swinir} & 0.4130 & 60.72 & 0.4503 & 51.51 \\
        RealESRGAN\cite{wang2021realesrgan} & 0.4483 & 62.99 & 0.4493 & 52.79 \\
        CDFormer\cite{liu2024cdformer} & 0.3826 & 28.37 & 0.3600 & 23.81 \\
        SinSR\cite{wang2024sinsr} & 0.5744 & 62.22 & 0.6342 & 54.40 \\
        DiffIR\cite{xia2023diffir} & 0.3963 & 60.30 & 0.4255 & 49.93 \\
        StableSR\cite{wang2024exploiting} & \underline{0.6008} & \underline{66.70} & \underline{0.6141} & \underline{56.89} \\
        Ours & \textbf{0.6453} & \textbf{66.90} & \textbf{0.6475} & \textbf{61.04} \\
        \bottomrule
    \end{tabular}
    }
    \label{tab:sr_combined_realsr_drealsr}
\end{table}

\section{Conclusion}
In this paper, we present a novel approach that capitalizes on the strengths of large-scale diffusion models to restore real-world low-quality images afflicted by unknown degradations while preserving the intuitive capabilities of text-based editing. Our method introduces Internal Image Detail Enhancement (IIDE) as a fine-tuning technique within the diffusion model, guiding the generative process to produce high-quality images from low-quality inputs while meticulously preserving intricate details. Our results affirm the effectiveness in providing object-level control over diversity while maintaining high visual consistency on multiple image restoration tasks, establishing its superiority over prior state-of-the-art models.

\clearpage

\appendix

\section{Appendix A: Compared Methods}
\subsection{Text-Guided Image Colorization} Since there have no existing works tackling the text-guided old photo restoration, to better demonstrate our method's text coloring capability, we synthesized a dataset with only grayscale degradation and used BLIP2 \cite{li2023blip} to generate text prompts, then separately compared with another two text-guided methods: UniColor \textsuperscript{{}}\footnote{Test on official model mscoco\_step259999.ckpt.} \cite{huang2022unicolor}, L-CoDe \textsuperscript{{}}\footnote{Test on our trained model using official codes.} \cite{weng2022lcode}. Specifically, we evaluate L-CoDe using an image size of $224 \times 224$, as the L-CoDe model exclusively supports this image scale.

\subsection{Old Photo Restoration} Since more recent methods Pik-Fix \cite{xu2022pikfix} have no publicly available models, we take two methods for evaluation: DeOldify \textsuperscript{{}}\footnote{Test on official model ColorizeStable\_gen.pth.} \cite{anctic2021deoldify} and BrOldPho \textsuperscript{{}}\footnote{Test on official repo: \url{https://github.com/microsoft/Bringing-Old-Photos-Back-to-Life}} \cite{wan2020bringing}. However, BrOldPho \cite{wan2020bringing} primarily concentrates on addressing the scratches on old photos, without color restoration. Conversely, DeOldify \cite{anctic2021deoldify} solely focuses on colorizing old photos, without attending to scratches. In order to facilitate a balanced comparison, we have amalgamated these two techniques into a novel processing method.

\begin{figure*}[!t]
  \centering
  \includegraphics[width=0.75\textwidth]{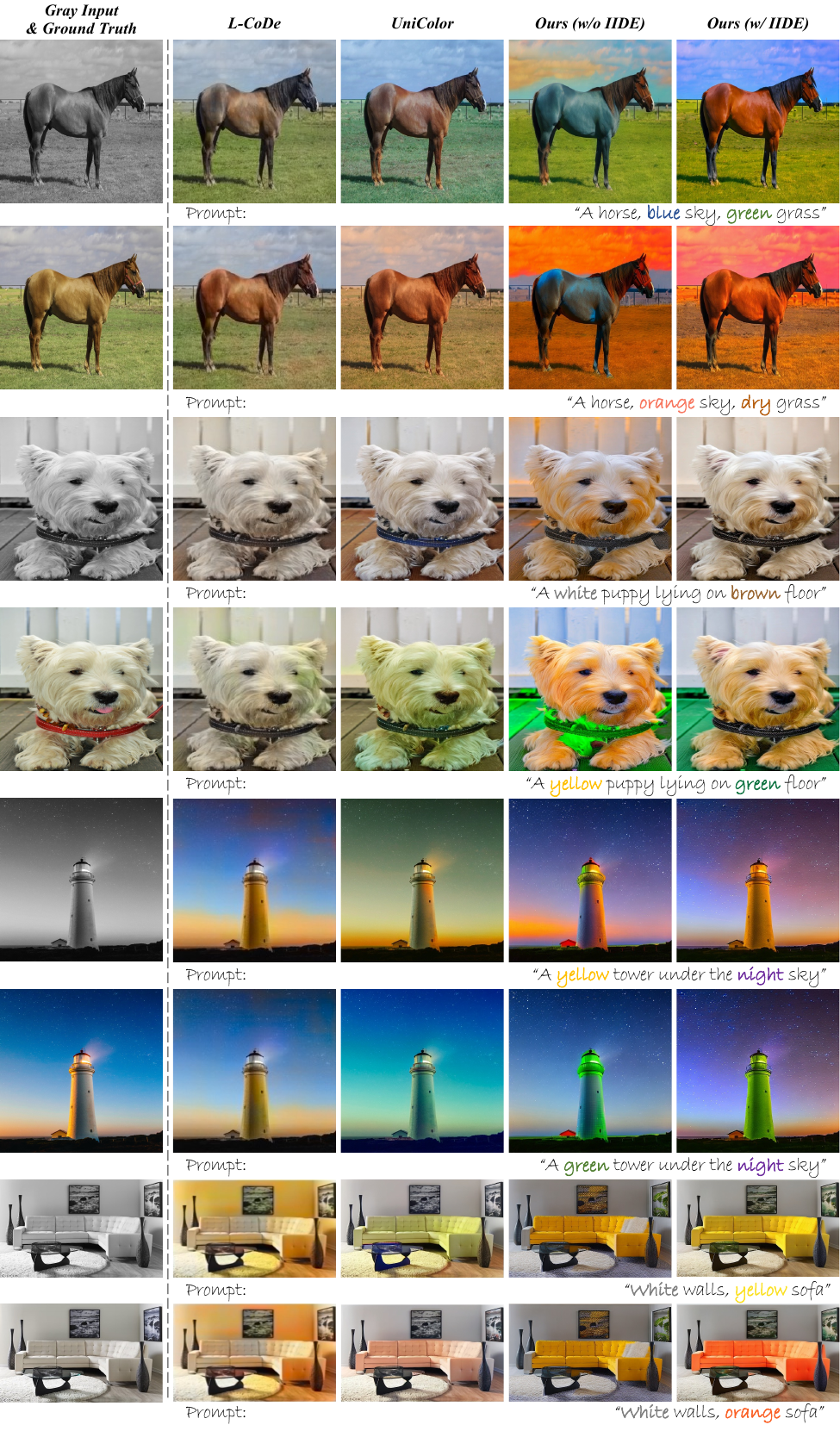}
  \caption{Visual comparison of text-guided image colorization methods. Zoom-in for better details.}
  \label{fig:image colorization with text prompt control}
\end{figure*}

\begin{figure*}[!t]
  \centering
  \includegraphics[width=0.8\textwidth]{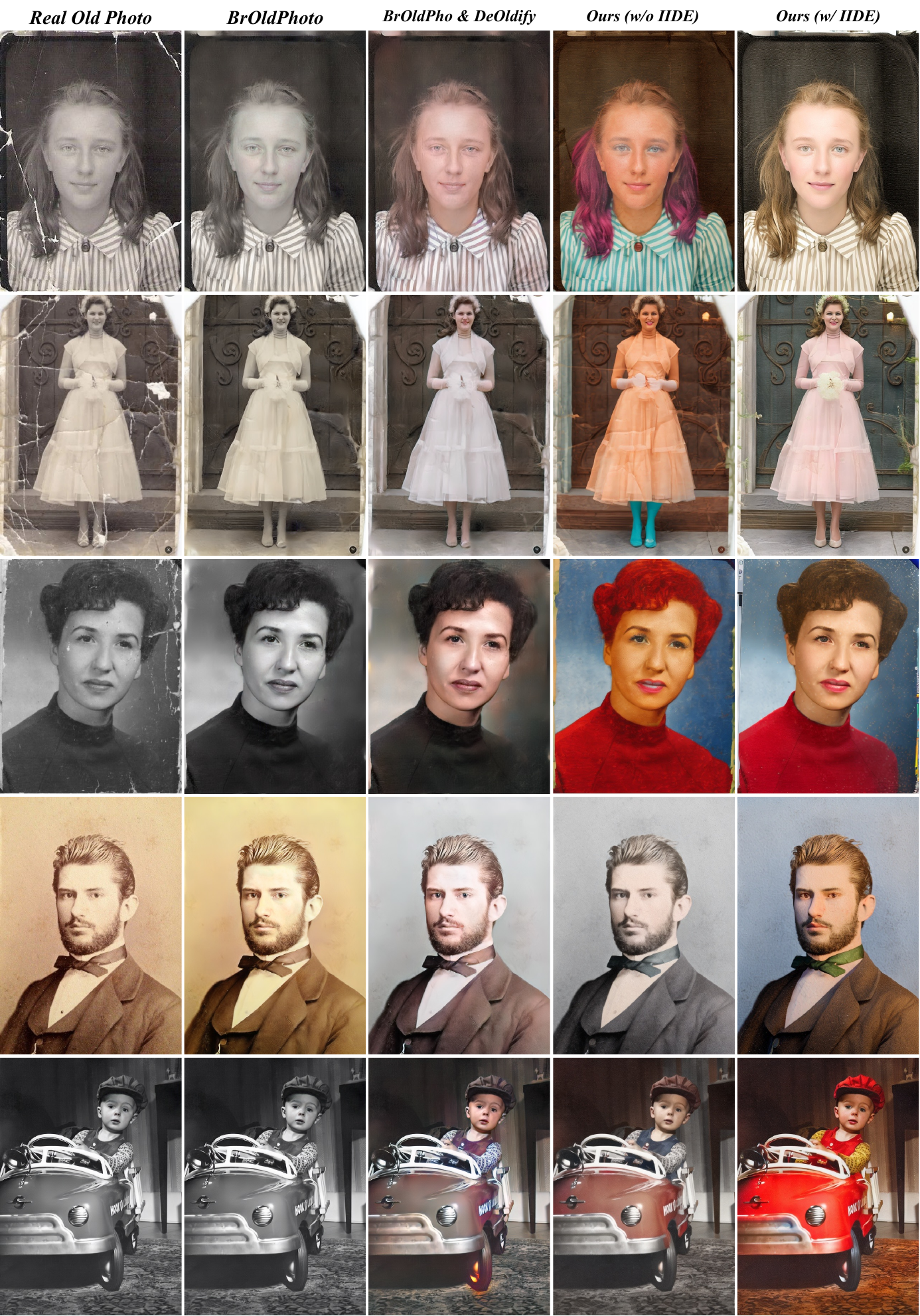}
  \caption{Visual comparison of real-world old photo restoration. Zoom-in for better details.}
  \label{fig:old photo restoration}
\end{figure*}

\begin{figure*}[!t]
  \centering
      \includegraphics[width=0.65\textwidth]{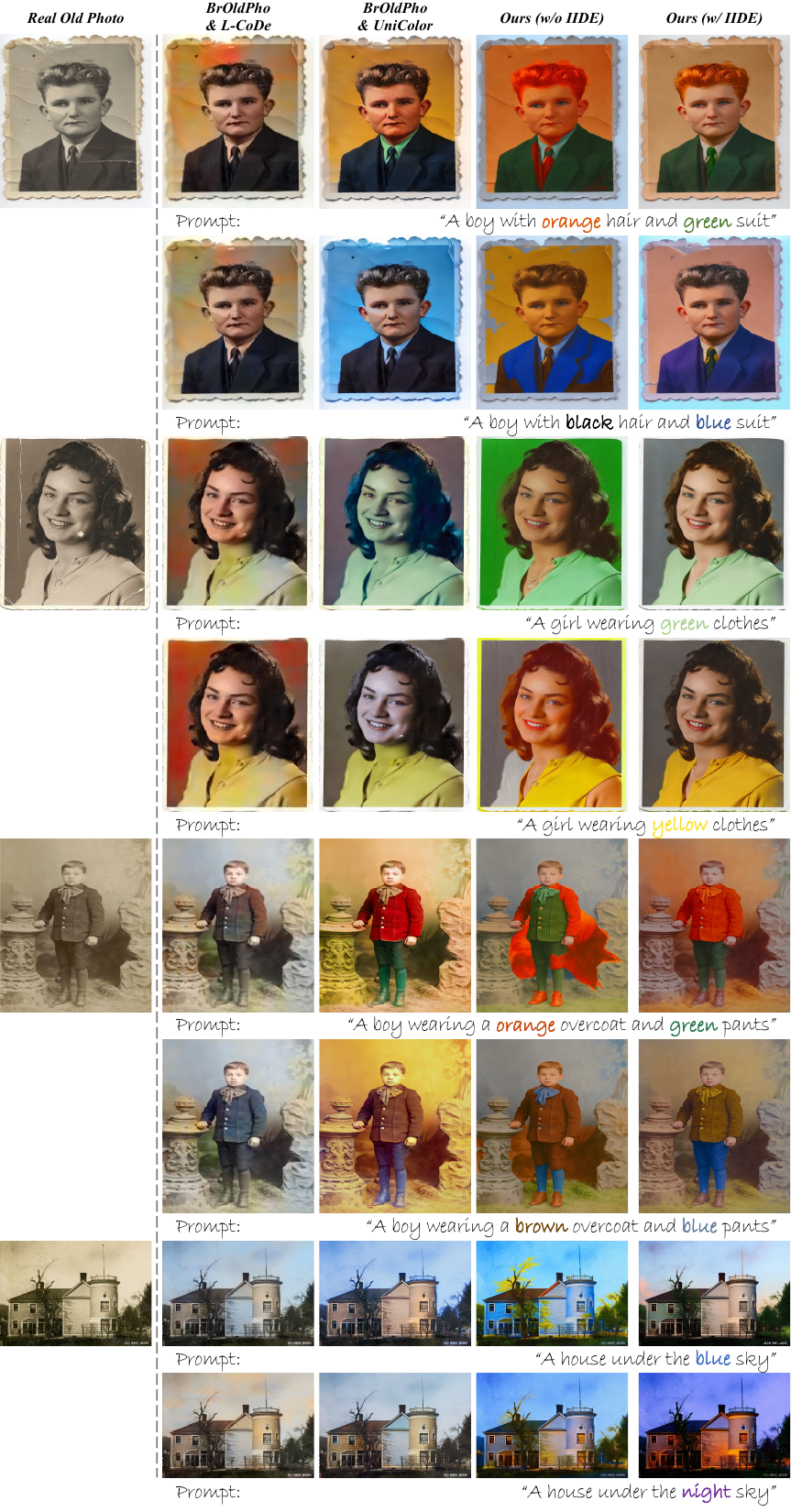}
  \caption{Visual comparison of text-guided real-world old photo restoration. Zoom-in for better details.}
  \label{fig:old photo restoration result with text prompt control}
\end{figure*}

\subsection{Image Super-Resolution} To verify the effectiveness of our approach, we compare our method with several state-of-the-art methods, i.e. SwinIR\cite{liang2021swinir}, RealESRGAN \textsuperscript{{}}\footnote{Test on official model RealESRGAN\_x4plus.pth.} \cite{wang2021realesrgan}, StableSR \textsuperscript{{}}\footnote{Test on official model stablesr\_000117.ckpt.} \cite{wang2023exploiting}, CDFormer \textsuperscript{{}}\footnote{Test on official model model\_1200.ckpt(cdformer\_x4\_bicubic\_iso).} \cite{liu2024cdformer}, SinSR\textsuperscript{{}}\footnote{Test on official model SinSR\_v1.pth.} \cite{wang2024sinsr} and DiffIR\textsuperscript{{}}\footnote{Test on official model RealworldSR-DiffIRS2-GANx4-V2.} \cite{xia2023diffir}. Specially, for methods based on diffusion models, low-resolution images are first resized from $128\times128$ to $512\times512$ before feeding into models.

\section{Appendix B: Evaluation on Text-Guided Image Colorization} 
Given the limited methods available for text-guided old photo restoration, we further assess the colorization capabilities of our model, which was originally developed for old photo restoration. As shown in Table \ref{tab:colorization}, although our model shows a slight decrease in PSNR and SSIM compared to UniColor, it outperforms in perceptual metrics such as CLIPIQA \cite{wang2023exploring} and MUSIQ \cite{ke2021musiq}. These results emphasize the effectiveness of our approach in producing perceptually superior colorized images, even when traditional fidelity metrics show a minor trade-off.

As illustrated in Fig. \ref{fig:image colorization with text prompt control}, our method produces more natural and text-accurate colors, while preserving fine details—highlighted by the green boxes on the license plate numbers. Despite the inherent challenges of colorization, the strong performance in perceptual metrics affirms the robustness of our model for text-guided image colorization.

\begin{figure*}[t]
  \centering
  \includegraphics[width=1.\textwidth]{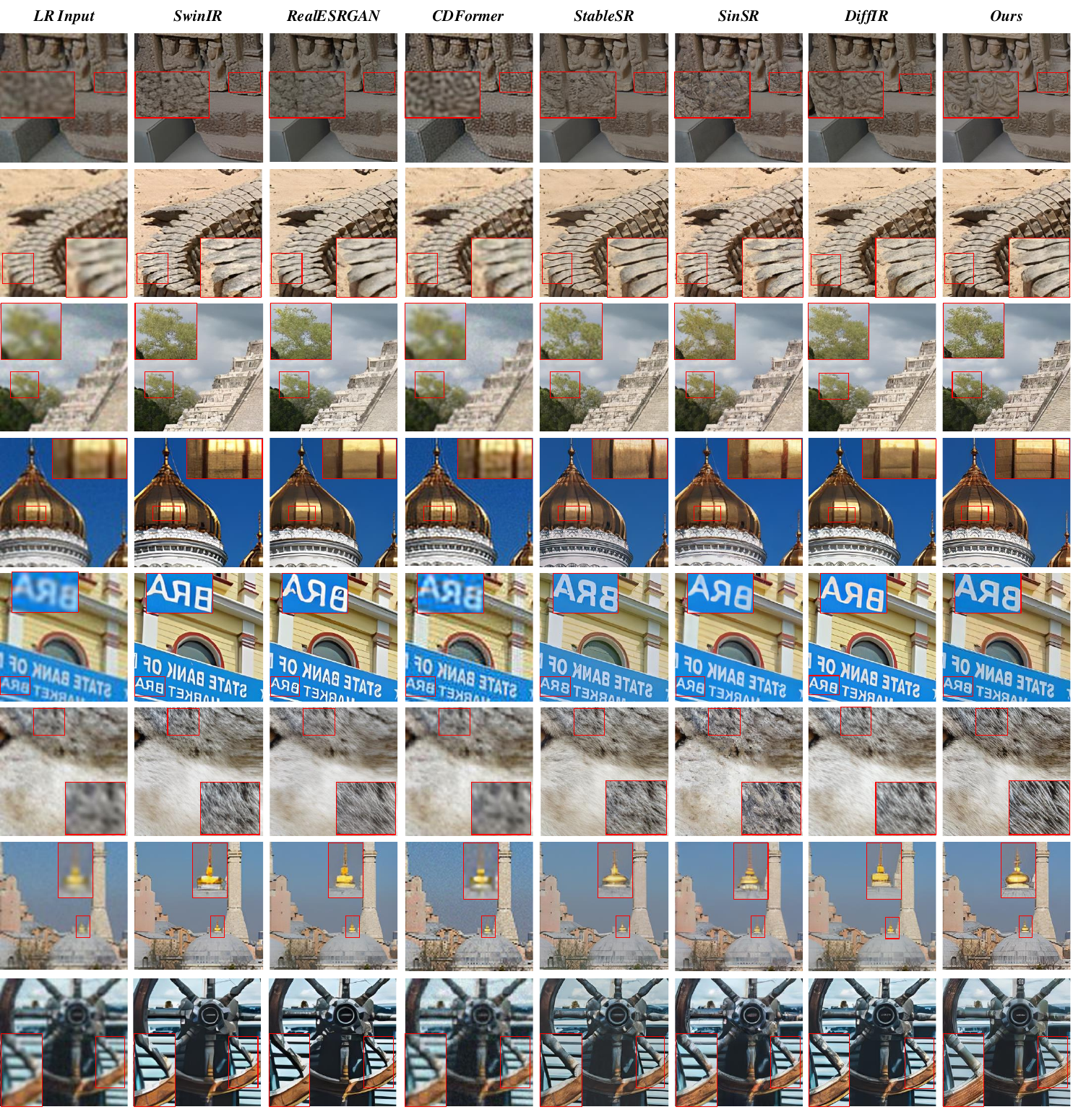}
  \caption{Visual comparison of image super-resolution methods on synthetic dataset. Zoom-in for better details.}
  \label{fig:div2k_realsr}
\end{figure*}

\begin{table}[!t]
    \centering
    \caption{Quantitative comparison results of Text-Guided Colorization on DIV2K.}
    \resizebox{\columnwidth}{!}{ 
    \small 
    \setlength{\tabcolsep}{4pt} 
    \renewcommand{\arraystretch}{1.2} 
    \begin{tabular}{c|cccccc}
        \toprule
        \multirow{2}{*}{\textbf{Methods}} & \multicolumn{6}{c}{\textbf{Metrics}} \\
        \cline{2-7}
        & \textbf{PSNR$\uparrow$} & \textbf{SSIM$\uparrow$} & \textbf{LPIPS$\downarrow$} & \textbf{FID$\downarrow$} & \textbf{CLIPIQA$\uparrow$} & \textbf{MUSIQ$\uparrow$} \\
        \midrule
        L-CoDe\cite{weng2022lcode} & \textbf{34.95} & \textbf{0.9671} & \underline{0.2356} & 49.06 & 0.5616 & 48.54 \\
        UniColor\cite{huang2022unicolor} & \underline{32.41} & \underline{0.9634} & \textbf{0.2027} & \underline{38.67} & 0.6924 & \underline{70.49} \\
        Ours (w/o IIDE) & 27.14 & 0.7971 & 0.4360 & 39.68 & \underline{0.7120} & 67.80 \\
        Ours (w/ IIDE) & 31.21 & 0.9543 & 0.2484 & \textbf{37.94} & \textbf{0.7490} & \textbf{74.37} \\
        \bottomrule
    \end{tabular}
    }
    \label{tab:colorization}
\end{table}

\begin{figure*}[t]
  \centering
  \includegraphics[width=1.\textwidth]{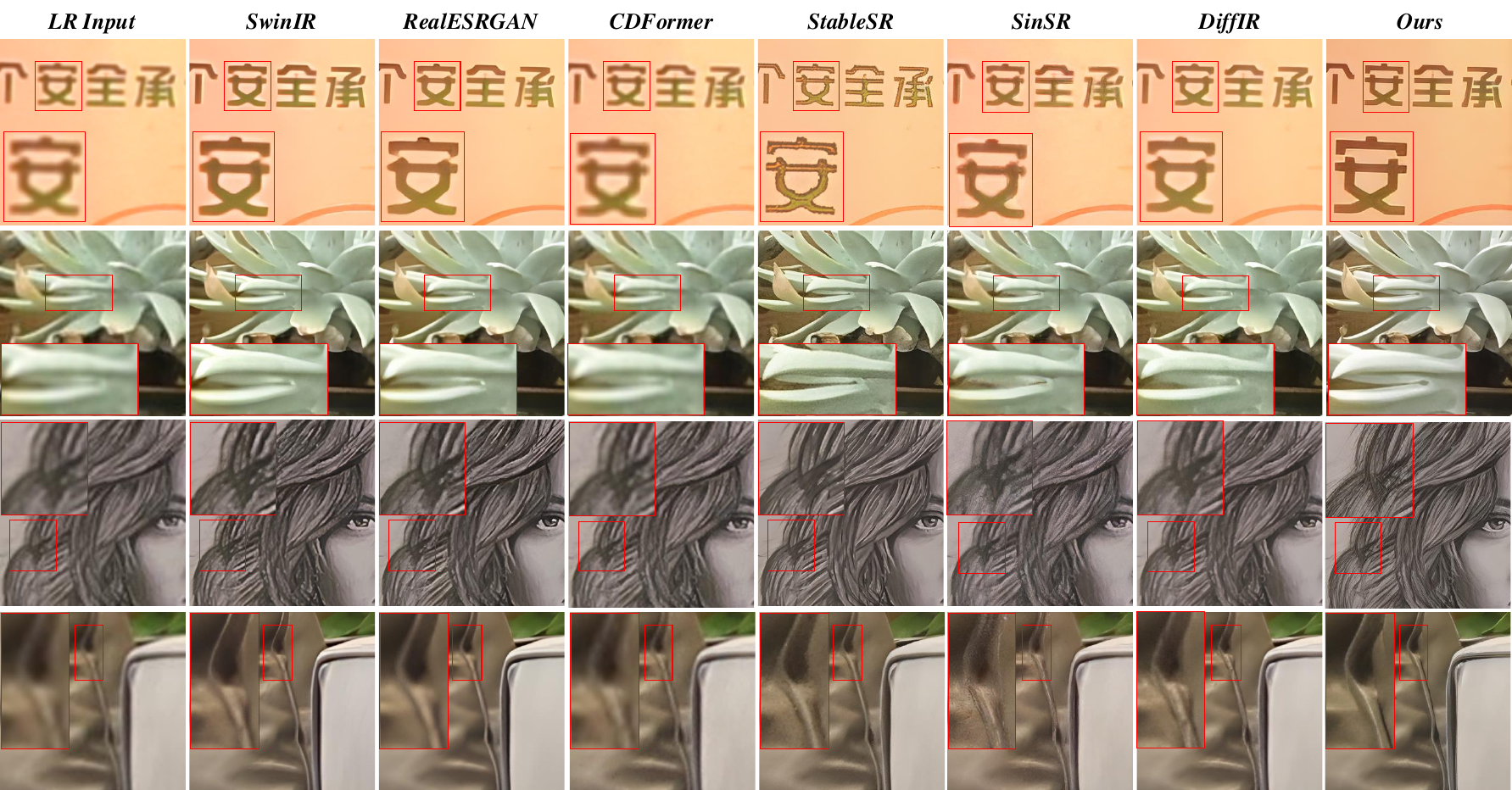}
  \caption{Visual comparison of image super-resolution methods on real-world datasets. Zoom-in for better details.}
  \label{fig:drealsr}
\end{figure*}

\subsection{Appendix C: Additional Visual Comparisons for Old Photo Restoration}

To comprehensively demonstrate the advantages of our approach, we present additional visual comparisons between automatic restoration (Fig. \ref{fig:old photo restoration}) and text-guided restoration (Fig. \ref{fig:old photo restoration result with text prompt control}) applied to real-world old photos sourced from the Internet. These comparisons clearly highlight the superior detail preservation and more vibrant color restoration achieved by our method across both tasks.

\subsection{Appendix D: Additional Visual Comparisons of Image Super-Resolution}
In this section, we provide further visual comparisons on both synthetic and real-world datasets to illustrate the effectiveness of our method in super-resolution tasks. Specifically, we present results on the synthetic DIV2K dataset \cite{agustsson2017ntire} in Fig. \ref{fig:div2k_realsr}, as well as on real-world datasets, including RealSR \cite{cai2019toward} and the DReal dataset \cite{wei2020component}, shown in Fig. \ref{fig:drealsr}. 

As demonstrated in these comparisons, our method not only recovers finer details but also produces more structurally coherent and visually natural images. In particular, the results on both the synthetic and real-world datasets highlight the superior clarity and fidelity of our restored images. Compared to prior methods, our approach consistently preserves edge sharpness, reduces artifacts, and enhances the perceptual quality, ensuring that the restored images not only appear sharper but also exhibit a more natural structure and texture. These improvements are especially evident in complex textures and fine structures, where our method demonstrates an ability to recover realistic details that are often lost in previous approaches.

\clearpage

\bibliographystyle{IEEEbib}
\bibliography{main}

\end{document}